\documentclass[journal]{IEEEtran}

\usepackage{graphicx,times,amsmath,amssymb,multirow,caption,subcaption,color}
\usepackage[LGR,T1]{fontenc}
\usepackage{algorithm,algcompatible}
\usepackage{cite}
\usepackage{color,soul}
\usepackage{blindtext}
\usepackage{csquotes}
\usepackage{graphicx}
\usepackage[legalpaper, margin=0.6in]{geometry}

\usepackage{fancyhdr}
\usepackage{textcomp}
\usepackage{scalerel}
\usepackage{csquotes}
\usepackage{subcaption}
\usepackage{multirow}
\usepackage{tikz}
\usetikzlibrary{svg.path}
\usepackage{tikz}
\usetikzlibrary{shapes,backgrounds}
\usepackage{verbatim}
\usepackage{url}
\usetikzlibrary{svg.path}

\definecolor{googleblue}{rgb}{0.259,0.522,0.957}
\definecolor{googlegreen}{rgb}{0.060,0.620,0.350}
\definecolor{googlered}{rgb}{0.859,0.267,0.220}
\definecolor{bleudefrance}{rgb}{0.19,0.55,0.91}

\definecolor{orcidlogocol}{HTML}{A6CE39}
\tikzset{
  orcidlogo/.pic={
    \fill[orcidlogocol] svg{M256,128c0,70.7-57.3,128-128,128C57.3,256,0,198.7,0,128C0,57.3,57.3,0,128,0C198.7,0,256,57.3,256,128z};
    \fill[white] svg{M86.3,186.2H70.9V79.1h15.4v48.4V186.2z}
                 svg{M108.9,79.1h41.6c39.6,0,57,28.3,57,53.6c0,27.5-21.5,53.6-56.8,53.6h-41.8V79.1z M124.3,172.4h24.5c34.9,0,42.9-26.5,42.9-39.7c0-21.5-13.7-39.7-43.7-39.7h-23.7V172.4z}
                 svg{M88.7,56.8c0,5.5-4.5,10.1-10.1,10.1c-5.6,0-10.1-4.6-10.1-10.1c0-5.6,4.5-10.1,10.1-10.1C84.2,46.7,88.7,51.3,88.7,56.8z};
  }
}

\newcommand\orcidicon[1]{\href{https://orcid.org/#1}{\mbox{\scalerel*{
\begin{tikzpicture}[yscale=-1,transform shape]
\pic{orcidlogo};
\end{tikzpicture}
}{|}}}}

\usepackage[bookmarks=false]{hyperref} 
\hypersetup{
    colorlinks=true,
}
\hypersetup{
  colorlinks = true,
  citecolor = black
}
\begin{document}
\raggedbottom

\twocolumn
\setcounter{page}{1}

\title{Advancing Frontiers in SLAM: A Survey of Symbolic Representation and Human-Machine Teaming in Environmental Mapping}

\author{
        \IEEEauthorblockN{Brandon C. Colelough,~$^{\orcidicon{0000-0001-8389-3403}}$}
        
        \IEEEauthorblockA{School of Engineering and Information Technology, University of New South Wales, Australia}
}

\maketitle

\textbf{Abstract} - \textbf{This survey paper presents a comprehensive overview of the latest advancements in the field of Simultaneous Localization and Mapping (SLAM) with a focus on the integration of symbolic representation of environment features. The paper synthesizes research trends in multi-agent systems (MAS) and human-machine teaming, highlighting their applications in both symbolic and sub-symbolic SLAM tasks. The survey emphasizes the evolution and significance of ontological designs and symbolic reasoning in creating sophisticated 2D and 3D maps of various environments. Central to this review is the exploration of different architectural approaches in SLAM, with a particular interest in the functionalities and applications of edge and control agent architectures in MAS settings. This study acknowledges the growing demand for enhanced human-machine collaboration in mapping tasks and examines how these collaborative efforts improve the accuracy and efficiency of environmental mapping}

\section{Introduction} \label{sec:intro}
As described in \cite{Ethical_AI}, there is an opportunity to use \enquote{artificial creativity to generate sensible narratives that summarise complex military situations for an operator that could enhance situational awareness (SA) where that operator is not already immersed in the low-level tactical situation}. Simultaneous Localisation and Mapping (SLAM) offers a solution to this capability gap and has great applicability over a range of military domains (see the appendix for use cases) and for which many systems have already demonstrated promising results \cite{Kimera_V2}, \cite{magic}, \cite{Hierarchical_prob_mapping}, \cite{TOSM} (see anything within the semantic-SLAM circle of figure~\ref{fig::venn_diagram} for more). The ethical use of AI-enabled systems within the military is predicated on the ability of the operator to effectively command such systems, as \enquote{command is a fundamentally human function that cannot be conducted by machines; it provides accountability for ethical decision making} \cite{ADF_Concepts}. The Defence Science and Technology Group list trust as a key facet required in any AI-enabled system to ensure an acceptable level of command is maintained over these systems \cite{DSTG}. The transparency in an AI-enabled system is the \enquote{operator’s awareness of an autonomous agent’s actions, decisions, behaviours and intentions} \cite{DSTG-2}. This system transparency is \enquote{an essential element to facilitate effective communication and collaboration} and as such is the \enquote{overarching concept} required for a human-machined teamed environment trust architecture \cite{swarm_ontology}. For further elaboration on the ethics of using AI-enabled systems within the Australian Defence Force, see the video listed here \cite{brandinio}. 
Recently, structured knowledge graphs known as ontologies have been applied to the field of simultaneous localisation and mapping (SLAM) (see figure~\ref{fig::venn_diagram} - Semantic SLAM, of note is \cite{TOSM}, \cite{Multi-Slam}, \cite{Kostavelis2015SemanticMF}, \cite{Robust_semantic_SLAM}, \cite{semantic_map_matching_2022}, \cite{robotics10040125}) allowing for semantic reasoning to generate more effective mapping techniques. Generally, two main orientations for contextual reasoning are employed, which are centralised and collaborative. Centralised observes agents transfer information to a back-end system for processing, whereas collaborative contains agents with the capability to locally process information and reason collaboratively \cite{predictive_edge}. Recent technological advances in artificial intelligence (AI) for visual reasoning have made it possible for edge agents to more effectively reason about their local environment \cite{edge_3d_reason} leading to a more effective integration of symbolic and sub-symbolic reasoning. A domain for which symbolic representation has not yet been fully realised, however, is the use of symbolic reasoning through ontological design to create contextual maps of an environment.


\def\firstcircle{(-1.5,1.5) circle (3cm)}
\def\secondcircle{(1.5,1.5cm) circle (3cm)}
\def\thirdcircle{(1.5,-1.5cm) circle (3cm)}
\def\fourthcircle{(-1.5,-1.5cm) circle (3cm)}

\begin{figure*}[ht]
    \centering
    \resizebox{\textwidth}{!}{%

\begin{tikzpicture}

    \begin{scope}[shift={(-4cm,0cm)}, fill opacity=0.5]
        \fill[red] \firstcircle;
        \fill[green] \secondcircle;
        \fill[blue] \thirdcircle;
        \fill[yellow] \fourthcircle;
        \draw \firstcircle node[above] {};
        \draw \secondcircle node [above] {};
        \draw \thirdcircle node [above] {};
         \draw \fourthcircle node [below] {};
    \end{scope}
\node at (-7,3) {\textbf{MAS}};

\node at (-1,3) {\textbf{Robotics}};

\node at (-7,-3) {\textbf{Simulation}};

\node at (-1,-3) {\textbf{Semantic }};
\node at (-1,-3.5) {\textbf{SLAM}};

\tiny

\node at (-4.5, 0) {\textbf{\cite{Multi-Slam}}};
\node at (-4, 0) {\textbf{\cite{TOSM}}};
\node at (-3.5, 0) {\textbf{\cite{Waibel2011RoboEarthA}}};
\node at (-4.2, -0.5) {\textbf{\cite{Kostavelis2015SemanticMF}}};

\node at (-8,2) {\textbf{\cite{AI_5G}}};
\node at (-7.5,2) {\textbf{\cite{angel}}};
\node at (-7,2) {\textbf{\cite{network_general}}};
\node at (-6.5,2) {\textbf{\cite{swarm_ontology}}};
\node at (-6,2) {\textbf{\cite{swarm_teaming_ontology}}};
\node at (-8, 1.5) {\textbf{\cite{footprints}}};
\node at (-7.5, 1.5) {\textbf{\cite{robot_swarms_search}}};
\node at (-7, 1.5) {\textbf{\cite{MAS_search}}};
\node at (-6.5, 1.5) {\textbf{\cite{DAI_Book}}};
\node at (-8, 2.5) {\textbf{\cite{State_of_the_art}}};
\node at (-7.5, 2.5) {\textbf{\cite{terrain_map}}};
\node at (-7, 2.5) {\textbf{\cite{group_mapping}}};
\node at (-6.5, 2.5) {\textbf{\cite{localization_3D_moving_target}}};
\node at (-6, 2.5) {\textbf{\cite{map_merge_review}}};
\node at (-6, 3) {\textbf{\cite{distributed_semantic_slam}}};
\node at (-7, 3.5) {\textbf{\cite{DOOR_Slam}}};
\node at (-6.5, 3.5) {\textbf{\cite{6D_map_match}}};
\node at (-6, 3.5) {\textbf{\cite{CCM_Slam}}};

\node at (-5, -1) {\textbf{\cite{Joo2019ARA}}};

\node at (-3.5, 1) {\textbf{\cite{crowd_source}}};

\node at (-7.5, 0) {\textbf{\cite{multi_uav_patrol}}};
\node at (-7, 0) {\textbf{\cite{Co-Slam}}};

\node at (-5, 2) {\textbf{\cite{MAS_graph_slam}}};
\node at (-4.5, 2) {\textbf{\cite{multi_robot_nav}}};
\node at (-4, 2) {\textbf{\cite{localization_review}}};
\node at (-3.5, 2) {\textbf{\cite{SLAM_Review}}};

\node at (-5, 1) {\textbf{\cite{abbass_hunjet_2022}}};

\node at (-1,2) {\textbf{\cite{holistic}}};
\node at (-1.5,2) {\textbf{\cite{Hierarchical_prob_mapping}}};
\node at (-2,2) {\textbf{\cite{sem_prob_reg}}};
\node at (-0.5,2) {\textbf{\cite{collab_semantic_understandng}}};
\node at (0,2) {\textbf{\cite{collab_perception_localization}}};
\node at (-0.5, 2.5) {\textbf{\cite{semantic_map_matching_2022}}};
\node at (0, 2.5) {\textbf{\cite{fusion_framework}}};
\node at (-1, 2.5) {\textbf{\cite{MAS_3D_Mapping}}};


\node at (-1.5, -1.5) {\textbf{\cite{DS_SLAM}}};
\node at (-1, -1.5) {\textbf{\cite{Bescs2018DynaSLAMTM}}};
\node at (-.5, -1.5) {\textbf{\cite{Ai2020DDLSLAMAR}}};
\node at (0, -1.5) {\textbf{\cite{Fan2020SemanticSW}}};


\node at (-2, -2) {\textbf{\cite{YOLO_Slam}}};
\node at (-1.5, -2) {\textbf{\cite{VC_Slam}}};
\node at (-1, -2) {\textbf{\cite{Yolo_Slam_2}}};
\node at (-0.5, -2) {\textbf{\cite{Blitz-Slam}}};
\node at (0, -2) {\textbf{\cite{yolo_slam_3}}};
\node at (-2, -2.5) {\textbf{\cite{Semantic_CRF}}};
\node at (-1.5, -2.5) {\textbf{\cite{incremental_segmentation}}};
\node at (-1, -2.5) {\textbf{\cite{distributed_semantic_slam}}};
\node at (-.5, -2.5) {\textbf{\cite{mask_fusion}}};
\node at (0, -2.5) {\textbf{\cite{Cui2019SOFSLAMAS}}};
\node at (-2.2, -3) {\textbf{\cite{Iqbal2020DataAA}}};
\node at (-2.5, -3.5) {\textbf{\cite{Zhang2019HierarchicalTM}}};
\node at (-2.7, -4) {\textbf{\cite{Guan2020ARS}}};
\node at (-3.2, -4) {\textbf{\cite{Zhang2019MaskRB}}};
\node at (-2.2, -4) {\textbf{\cite{QuadricSlam}}};
\node at (-1.7, -4) {\textbf{\cite{Cascianelli2017RobustVS}}};
\node at (-1.2, -4) {\textbf{\cite{vis_landmarks}}};

\node at (-5, -2) {\textbf{\cite{SimVOIDS}}};
\node at (-4.5, -2) {\textbf{\cite{data_assosc}}};
\node at (-4, -2) {\textbf{\cite{Kimera}}};
\node at (-3.5, -2) {\textbf{\cite{Kimera_V2}}};
\node at (-3, -2) {\textbf{\cite{Slam++}}};

\node at (-8,-2) {\textbf{\cite{Multi-Store}}};
\node at (-7.5, -2) {\textbf{\cite{RuizSarmiento2015ExploitingSK}}};
\node at (-7, -2) {\textbf{\cite{KnowRobSim}}};
\node at (-6.5, -2) {\textbf{\cite{auto_ontology}}};
\node at (-6, -2) {\textbf{\cite{Webots}}};'
\node at (-8, -2.5) {\textbf{\cite{Koenig2004DesignAU}}};

\node at (-0.5,0) {\textbf{\cite{Semantic_SLAM}}};
\node at (-1,0) {\textbf{\cite{Robust_semantic_SLAM}}};
\node at (-1.5,0) {\textbf{\cite{3D_semantic_SLAM}}};
\node at (-2,0) {\textbf{\cite{DONG2022153}}};
\node at (-2.5, 0) {\textbf{\cite{submap_matching}}};


\node at (-0.7, -.5) {\textbf{\cite{semantic_map_build}}};
\node at (-1.2, -0.5) {\textbf{\cite{Hierarchical_SLAM}}};
\node at (-1.7, -0.5) {\textbf{\cite{Nchter2008TowardsSM}}};


\node at (-0.7, 0.5) {\textbf{\cite{Xiao2019DynamicSLAMSM}}};
\node at (-1.2, 0.5) {\textbf{\cite{volumetric_slam}}};
\node at (-1.7, 0.5) {\textbf{\cite{Prob_sem_slam_orig}}};
\node at (-2.2, 0.5) {\textbf{\cite{multi_sem_slam}}};


\node at (-1.2, 1) {\textbf{\cite{Cube_Slam}}};
\node at (-1.7, 1) {\textbf{\cite{dynamic_industrial_slam}}};
\node at (-2.2, 1) {\textbf{\cite{Wang2020RobustLC}}};

\node at (-3.5, -1) {\textbf{\cite{SO-Slam}}};
\node at (-3, -1) {\textbf{\cite{De-Slam}}};
\node at (-2.9, -0.5) {\textbf{\cite{Planar}}};

\normalsize
\node at (-6,-5){\textbf{Ontologies for SLAM}};
\tiny
\draw (0,-5.5) -- (-8,-5.5) -- (-8,-7) -- (0,-7) -- (0,-5.5);
\node at (-7.5,-6)  {\textbf{\cite{robotics10040125}}};
\node at (-7,-6) {\textbf{\cite{Tenorth2013KnowRobAK}}};
\node at (-6.5,-6) {\textbf{\cite{Know_rob_map}}};
\node at (-6, -6) {\textbf{\cite{Schlenoff2005ARO}}};
\node at (-5.5, -6) {\textbf{\cite{Mozos2007FromLT}}};
\node at (-5, -6) {\textbf{\cite{Suh2007OntologybasedMR}}};
\node at (-4.5, -6) {\textbf{\cite{Eid2007AUO}}};
\node at (-4, -6) {\textbf{\cite{Belouaer2010OntologyBS}}};
\node at (-3.5, -6) {\textbf{\cite{Cashmore2015ROSPlanPI}}};
\node at (-3, -6) {\textbf{\cite{Silva2013TowardsAC}}};
\node at (-2.5, -6) {\textbf{\cite{Chang2020OntologybasedKM}}};
\node at (-2,-6)  {\textbf{\cite{Lim2011OntologyBasedUR}}};
\node at (-1.5,-6) {\textbf{\cite{Dhouib2011ControlAC}}};
\node at (-1, -6) {\textbf{\cite{Pronobis2011MultimodalSM}}};
\node at (-0.5,-6) {\textbf{\cite{Wang2011ObjectSM}}};


\node at (-7.5, -6.5) {\textbf{\cite{Riegen2012CombiningQS}}};
\node at (-7, -6.5) {\textbf{\cite{Paull2012TowardsAO}}};
\node at (-6.5, -6.5) {\textbf{\cite{Li2013MultilayeredMB}}};
\node at (-6, -6.5) {\textbf{\cite{Carbonera2013DefiningPI}}};
\node at (-5.5, -6.5) {\textbf{\cite{Wu2014SpatialSH}}};
\node at (-5, -6.5) {\textbf{\cite{Riazuelo2015RoboEarthSM2}}};
\node at (-4.5, -6.5) {\textbf{\cite{Burroughes2016OntologyBasedSG}}};

\node at (-4,-6.5)  {\textbf{\cite{Ramos2018OntologyBD}}};
\node at (-3.5,-6.5) {\textbf{\cite{Deeken2018GroundingSM}}};
\node at (-3,-6.5) {\textbf{\cite{Sun2019HighLevelSD}}};
\node at (-2.5, -6.5) {\textbf{\cite{Crespo2020SemanticIF}}};
\node at (-2, -6.5) {\textbf{\cite{Joo2020AutonomousNF}}};
\node at (-1.5, -6.5) {\textbf{\cite{Karimi2021AnOA}}};
\node at (-1, -6.5) {\textbf{\cite{Shchekotov2021TheOD}}};
\node at (-0.5, -6.5) {\textbf{\cite{9091567}}};

\end{tikzpicture}
}
\caption{A literature review of existing architectures surrounding the concept of the proposed SYMBO-SLAM architecture was conducted. This literature review focused using symbolic reasoning through ontological design to create contextual maps of an environment. The Venn diagram in the centre links the majority of concepts required for the implementation of the proposed system in simulation and on hardware. The box below lists the papers directly applicable to the use of symbolic reasoning in the SLAM domain}
\label{fig::venn_diagram}
\end{figure*}


\section{Literature Review} \label{sec::lit_review}
A systematic literature review (SLR) is the process of \enquote{identifying, evaluating and interpreting all available research relevant to a particular research question or topic area or
phenomenon of interest} \cite{Kitchenham07guidelinesfor}. To prove the capability gap for the proposed system existed, a systematic literature review surrounding the core concepts of the thesis main and sub questions was conducted in accordance with the PRISMA review standard \cite{prisma} and with consideration to the software engineering SLR guidelines listed in \cite{Kitchenham07guidelinesfor}.

\subsection{Search Process} \label{sec::search_process}
A manual search process was conducted to identify published peer-reviewed articles. This process utilised the following databases:

\begin{enumerate}
    \item IEEE Explore:  \url{http://ieeexplore.iee.org}
    \item Scopus: \url{https://www.scopus.com/}
    \item Semantic Scholar: \url{https://www.semanticscholar.org/}
    \item Google Scholar: \url{http://scholar.google.com}
    \item arxiv online library: \url{https://arxiv.org/} 
    \item Science Direct: \url{www.sciencedirect.com}
    \item Multidisciplinary Digital Publishing Institute (MDPI): \url{https://www.mdpi.com/}
    \item The Association for Computing Machinery (ACM) digital: \url{ http://portal.acm.org/portal.cfm}
    \item SpringerLink Library: \url{www.springerlink.com}
\end{enumerate}


\begin{figure*}[ht]
    \centering
    \resizebox{\textwidth}{!}{
   
    \begin{tabular}{ |p{3cm}||p{3cm}|p{3cm}|p{3cm}|p{3cm}|p{3cm}|p{3cm}|}
     \hline
     \multicolumn{7}{|c|}{Literature Results from Database queries} \\
     \hline
    Database & multi-agent SLAM & semantic SLAM & symbolic SLAM & ontology SLAM &  human swarm team & human MAS team\\
     \hline
     IEEE   & 28 & 23 & 4 & 6 & 9 & 1\\
     Scopus & 33 & 30 & 35 & 23 & 37 & 8\\
     Semantic Scholar & 30 & 30 & 37 & 35 & 35 & 31\\
     Google Scholar & 38 & 30 & 2 & 2 & 50 & 0\\
     ArXiv & 11 & 30 & 1 & 0 & 17 & 3\\
     Science Direct & 4 & 26 & 0 & 0 & 4 & 0\\
     MDPI   & 7 & 30 & 0 & 2 & 4 & 0\\
     ACM  & 2 & 9 & 0 & 0 & 1 & 0\\
     Springer & 2 & 18 & 0 & 0 & 4 & 0\\
     \hline
    \end{tabular}
    
    }

\captionof{table}[Hits_Results]{The 6 search terms were queried through the 9 databases as described above. The number of pieces of literature returned from each query is shown in the table above. Note also that results that returned above 30 literature pieces were further filtered by year published for reduction}
\label{fig::hits_results}
\end{figure*}


In addition to the papers selected from these databases, the bibliographies of papers selected through the SLR process were scanned for additional papers (snowballing). The search terms used to conduct the SLR were:

\begin{enumerate}
    \item multi-agent AND SLAM 
    \item semantic AND SLAM
    \item symbolic AND SLAM
    \item ontology AND SLAM
    \item human AND swarm AND teaming
    \item human AND MAS AND teaming
\end{enumerate}

\subsection{Inclusion / Exclusion Criteria and Quality Assessment} \label{sec::inc_exc}
Articles were included if they were: 
\newline
\begin{itemize}
    \item Published in a conference or journal or were book chapters 
    \item Articles published from 2010 to 2022
    \item Full text articles 
\end{itemize}

Articles were excluded if they were:
\begin{itemize}
  \item Not written in English
  \item Similar articles from different databases 
  \item Articles not relating to the main concept areas being SLAM or MAS or symbolic reasoning 
\end{itemize}

\subsection{Paper Selection} \label{sec::Paper_Selection}
Five focal areas directly applicable to the main research question and the mechanisms to achieve it were formulated. These include:

\begin{enumerate}
    \item Semantic SLAM 
    \item ontologies for SLAM 
    \item Multi-Agent Systems 
    \item Simulation
    \item Robotics 
\end{enumerate}

Nine additional areas were also considered during the literature review process that formulated key areas of research applicable to this project, which included:

\begin{enumerate}
    \item ontologies and Symbolic Reasoning 
    \item Map Matching
    \item Localisation 
    \item 2D Mapping Techniques
    \item 3D Mapping Techniques
    \item Agent Search and Edge Reasoning
    \item Path / Mission Planning
    \item Tracking and Spatio-Temporal Reasoning 
    \item Semantic Segmentation 
\end{enumerate}

The papers obtained from the manual search were then evaluated against these areas for relevance. The title, abstract and conclusion were considered to determine whether the paper related to the main concept areas for the project as listed above. 117 papers were then selected that had direct relevance to the five focal areas for the research project. These papers were then reviewed in depth and categorised to be listed in the Venn diagram shown in figure ~\ref{fig::venn_diagram}. This diagram aims to illustrate the specific area of relevance for each article reviewed in the systematic literature review, and links the majority of concepts required for the implementation of the proposed system in simulation and on hardware. These 117 papers were then also sub-categorised further to show relevance to each of the nine additional areas of relevance listed above. During this process, the snowball method was utilised to highlight an additional 78 pieces of literature from the references in the original set that had direct relevance to subject areas surrounding  the use of symbolic-reasoning through ontological design to create contextual maps of an environment on a MAS with SLAM. The 195 articles in total were then placed in the categorised boxes surrounding the Venn diagram. These sub-categories aimed to support the underlying theory of the proposed system. See the appendix - figure 23 for the Venn diagram figure of the full literature review conducted.

\subsection{Agent search and path planning} \label{sec::search}
It is natural to assume that the optimal path will be the main area of interest for search and detection problems. This is often not the case, and instead the distribution effort of searchers (especially on a MAS) is the object of interest for effective search strategies. As Washburn describes in his book\cite{washburn_2014}, there are two good reasons for this:
\begin{enumerate}
    \item Most searchers would not benefit from knowing the optimal path unless it happened to be easy to follow
    \item a “path” is not a convenient mathematical object in most cases
\end{enumerate}
As such, Washburn poses that for a fixed region, a random search pattern will yield a detection rate in less time than that of an exhaustive search. This decrease in detection time does however come at the cost of area coverage. With a strategy attempting to cover the searchable fraction $ A' =VWt $ of area A (where V is the speed and, W is the sweep width of a region and t is the time taken), an exhaustive search will cover $A'/A$ of that area (or else all of that area). This is not the case however for a random search strategy. Washburn uses the analogy of a searcher randomly dropping confetti within the area A to demonstrate this, stating that a random search is \enquote{effectively confetti casting, and it achieves a lower detection probability than the exhaustive search because of the wasteful overlap of one piece of confetti on another}. 
 Resource management in a constrained networked environment is vital for distributed multi-agent based systems such as the one proposed in this project. Al-Asfoor et al. \cite{MAS_search} propose  search algorithms based on a multi-agent model for dynamic and heterogeneous networks. This search algorithm makes use of conventional random walk algorithm with a semantic inspired resource search paradigm to direct search messages based on the semantic closeness of a target object. Duncan et al. \cite{robot_swarms_search} further introduce coverage and targeting Levy strategies that are biologically inspired and have been adapted to robotic swarm systems, which take their basis again in random search strategies.

\subsection{Feature extraction} \label{sec::feat_ext}
The term feature extraction in machine learning typically refers to the process of dimensionality reduction by which an initial set of raw data is reduced to a more manageable group for processing. This process aims to find the most compacted and informative set of features (distinct patterns) to enhance the efficiency of a classifier \cite{SUBASI2019193}. We define feature extraction as \enquote{the process of extracting key features (landmarks) from the environment through visual processing,} which is specific to the setting proposed. Current object detection architectures such as EfficientDet \cite{Tan2020EfficientDetSA}, ASSFF \cite{Zhang2020BridgingTG}, CenterMask \cite{Lee2020CenterMaskRA} and ATSS \cite{Zhang2020BridgingTG} offer an average precision between 43\% and 50\% depending on a multitude of variables such as speed and hardware implementation. Presently, the well recognised YOLO algorithm  dominates the field of object detection with its fourth implementation \cite{Bochkovskiy2020YOLOv4OS} (note that the fourth version is the latest from the original creator and with a published paper) and comes packaged with pre-trained weights and an enormity of training data-sets (ImageNet, VOC, Kaggle etc).

\subsection{Place Recognition} \label{sec::place_rec}
Place recognition is the ability to identify a feature regardless of change in pose (viewpoint) or other factors of appearance within a global map. This technique is commonly used in literature to enable feature localisation on a global co-ordinate system. In their 2009 survey, Williams et al. \cite{WILLIAMS20091188} compared the state-of-the-art place recognition techniques used for loop closure technology. They concluded that, due to the sparsity of information, map matching techniques under-performed when compared to image matching techniques. ORB-SLAM \cite{orbslam} \cite{OrbSlam2} is a well known visual-SLAM architecture which utilises image matching techniques for loop closure detection. The embedded place recognition module was based on the bag-of-words place recognition technique \cite{BOW} and implements a covisibility graph of features known as ORBs -Orientated FAST (key point detector architecture \cite{Rosten2006MachineLF}) and Rotated BRIEF (descriptor architecture \cite{Calonder2010BRIEFBR}). These features are binary features invariant to rotation and scale \cite{Rublee2011ORBAE} used to simultaneously conduct tracking, local mapping and loop closing. This system dominated the V-SLAM field for a period post its release in 2015 prior to being superseded by more updated machine-learning place recognition techniques. A more recent study from 2021  \cite{DBLP:journals/corr/abs-2106-10458} compared the deep learning approaches to place recognition, which utilised methods on the spectrum of supervised to unsupervised learning categories. End-to-end frameworks used to address a domain translation problem for place recognition were also explored in this survey, for which NetVlad \cite{NetVlad} was found to have had the largest impact. In the domain of supervised learning, three approaches were explored, being holistic, landmark and region based. From this, landmark based supervised training was found to effectively solve for appearance change, perceptual aliasing and viewpoint changing. These techniques (as demonstrated in \cite{doi:10.1177/0278364919839761}, \cite{DBLP:journals/corr/abs-1804-05526}, \cite{8653820},  \cite{9564866}, \cite{covnet_landmarks}) employ feature extractors (Generally CNN's or similar)  to develop semantic-based feature labels used to identify potential landmarks in a visual feed (see localisation on figure 23 for more). Several systems (such as \cite{Kimera_V2}, \cite{DOOR_Slam}, \cite{nonparametric_slam} to name a few) utilise these extracted land-marks to generate a pose-graph of key features within an environment. As such, the area of pose graph optimisation is a relevant field of study for land-mark based SLAM (see \cite{PGO1} for a good overview on this).

\subsection{Multi-Agent System} \label{sec::MAS}
An agent within a MAS has characteristics of self-sufficiency, social capacity and reactivity\cite{DAI}. A multi-agent system consists of any number of agents greater than two. Processing on a Distributed Artificial Intelligence (DAI) network can be done either through Mobile Cloud Computing (MCC) or Multi-Access Edge Computing (MEC). The MCC approach observes a system where the majority of the processing requirements is completed at the back end of the system, whereas MEC employs an architecture where the majority of the processing is done at the edge \cite{AI_5G} \cite{angel}. Similarly, two main orientations for the contextual reasoning of an environment through edge agents are generally given as centralised contextual reasoning where agents transfer information to a back-end system for processing and collaborative where agents locally process information and reason in a collaborative manner \cite{predictive_edge} \cite{network_general}.

\subsection{Map Matching} \label{sec::map_matching}
All SLAM algorithms require some form of map merging technique to be applied to enable the continuous updating of the known environment currently being mapped. As described in \cite{WILLIAMS20091188}, three main approaches were traditionally used, being map-to-map, image-to-image and image-to-map. These approaches generally took the form of frame-to-frame, single rigid body transformations to relate reference frames in different maps. Early implementations of SLAM such as \cite{magic} would generally use spatial matching between sub-maps to achieve the map matching process. This was done by representing the 6D voxels of an RGB-D sensor or point-cloud system as a Gaussian Distribution that can subsequently be integrated together as a map-merging technique. \cite{map_merge_review}  details a networked solution for feature merging through map alignment and data association for applications with SLAM solutions on a MAS. Similar map matching techniques are still being advanced today as shown in \cite{semantic_map_matching_2022} where semantic labelling is integrated with geometric map matching to generate a more effective SLAM result for loop closure. SLAM techniques that utilised key frame solutions allowed for map matching to occur as a data insertion into a graph rather than the geometric matching techniques used previously as seen here \cite{OrbSlam2} \cite{Kimera_V2}, \cite{nonparametric_slam} with pose-graphs. Of interest was a technique proposed in \cite{State_of_the_art} that allowed for cross-platform map merging between platforms that did not share the same sub-mapping techniques, which proposed the use of semantic representation for common understanding. For a more in depth explanation on traditional map matching methods, see \cite{DBLP:journals/corr/abs-2106-10458} sub-paragraph 3 – belief generation. Recently, techniques for map matching utilising landmark features (as described above) have seen success. \cite{8653820} demonstrates a ConvNet landmark-based visual place recognition system that utilises sequence search and hashing-based landmark indexing, which greatly increased the efficiency of the map-matching process. \cite{doi:10.1177/0278364919839761} is another place recognition architecture that employs semantic understanding of landmark features for an effective map matching. This system employed  CNN-based key-point matching, which utilised semantic filtering and dense descriptor weighting to allow for a place search procedure leading to a candidate match selection function.

\subsection{Ontology and Symbolic Reasoning} \label{sec::ontology}
Symbolic AI is the term used for a number of related AI methods that attempt to reason
about problems using high-level human understandable representations (symbols) \cite{DSTG}. For contrast, sub-symbolic AI systems (such as artificial neural networks, deep learning etc.) attempt to learn operations from (often large) data sources to map the variables in a problem space.

\[ f: X \xrightarrow[]{} Y \]

A key takeaway from this is that (even at high levels of complexity), sub-symbolic methods for machine learning determine some output \(Y\) through a learned function \(f\) (e.g., weights in a CNN) without having a clearly explainable or transparent process for obtaining said outputs. As such, some form of symbolic reasoning is required to generate a level of trust in the production of useful ground maps capable of summarising complex situations. This would enable their authenticity and as such enhance the situational awareness of an operator that is not already immersed in the low-level tactical situation \cite{DSTG}. As described in \cite{jean-baptiste_2021}, an ontology is a set of entities, which can be classes, properties, or individuals. ontologies may be used to  standardise the knowledge base of a specific domain and allow readability by both human and machines, as has been demonstrated in \cite{swarm_teaming_ontology}. Ontologies are designed to represent complex knowledge sets about things and the relations between them. As described in chapter 1 subsections 2-3 of \cite{onto_engineering}, the common components of an ontology are its individuals (or instances, I), classes (or concepts, C), attributes (a) and the relationships (R) linking them. Restrictions and rules can be placed upon these components to assert some knowledge about them. Axioms (A) are developed within the ontology to generate assertions to describe the overall theory of the ontology for its application domain. These assertions are provided in a logical form, such as for example:

\[ A \implies B \]
\[ A \text{  exists} \] 
\[ \therefore B \]

And hence, an ontology can be described (as it commonly is in literature) as a five-tuple \cite{Onto4MAT}:

\[ O = <C,R,a,I,A> \]

Transparency refers to an operator's awareness of an autonomous agent's actions, decisions, behaviours and intention \cite{Ethical_AI}. This system transparency is the overarching concept required to enable a trust architecture within the human-machine teamed environment. This transparency can be achieved through multiple methods such as interpretability, explainability and predictability \cite{swarm_ontology}. Utilising these ontologies provides a method to share semantic knowledge, promoting bi-directional transparency between humans and artificial agents \cite{Onto4MAT} through the three key facets listed above.  Therefore, the sub-set of symbolic AI that is ontological design offers enables explainability, interpretability and predictability within a system which is required to achieve transparency in an artificially intelligent system and hence enable the teamed operator to trust a swarm of robots to complete a cooperative function \cite{doi:10.1177/0018720819879273}. ontologies have been applied extensively in many fields to incorporate expert level domain knowledge into symbolic solutions. A corpus for ontology and semantics was proposed in \cite{VC_Slam} that leverages the textual data documentations for semantic labelling and modelling from 101 separate data sets, which presents a great opportunity for researchers to create more thorough ontological-domain representative models. Utilising these hierarchical ontologies for guided learning in sub-symbolic systems is a concept that is explored by Campbell \cite{abbass_hunjet_2022} (Chapter 6). This application enables a sub-symbolic system to reason on abstract concepts and reduce the dimensionality of a problem space (through partitioning) by applying  prior knowledge to a learning system. Such a  system is described by Hepworth in \cite{abbass_hunjet_2022} (Chapter 7). Here, a hybrid approach to activity recognition is detailed that fuses both data- and ontology-driven approaches to the activity recognition problem space in the machine learning domain. A similar approach to symbolic reasoning through the use of a knowledge graph to allow for a more diverse object detection system is shown in \cite{semantic_reason_few_shot} and a similar use of semantics to broaden the knowledge depth of scenes for VQA tasks is shown in \cite{explicit_knowledge}. 
The RoboEarth framework presented in \cite{RoboEarth_3}, \cite{RoboEarth_4} proposed a system that inherently integrates a knowledge base with visual SLAM, which in turn allowed for  more accurate representation of the environment and recognition of the objects found within. The semantic modelling framework Triplet Ontological Semantic Model (TOSM) proposed in \cite{TOSM} utilised short and long term memory with a static ontology to create of an on-demand ontological knowledge graph for the representation of an environment in the ontological domain (see figure ~\ref{fig::venn_diagram} - ontologies for SLAM for more). 

The HST-3 architecture as described in \cite{swarm_ontology} was developed to enable transparency for the application of human-machine teaming. The authors here specify that interoperability, explain-ability and predictability are the tenets required to enable transparency in an AI system which allowed for the operator to further trust the system as a whole. The decentralised nature of reasoning conducted on a MAS is described here as a main challenge in explain-ability (and hence transparency) of a system. The accumulation of varying agent experiences leads to a heterogeneous knowledge set, which hinders convergence to a central solution. To alleviate this issue, control agents are utilised to ensure the responsibility of explanation rests on only a few agents. Similarly, the proposed SYMBO-SLAM architecture aims to use the control agent architecture to alleviate this issue. The HST-GO architecture built upon the HST-3 framework described in \cite{swarm_teaming_ontology} aims to achieve transparency in the human-machine interaction space through bi-directional information flow. The tenets of lexicon, syntax and semantics are used here to describe the shared language between the individuals of the system and further enable communication in the human-swarm teaming environment. The most advanced ontology designed specifically for SLAM tasks is presented in \cite{robotics10040125}. This ontology suite standardises the SLAM problem set, combining data from a range of ontology sets to achieve superiority at the domain knowledge, lexical and structural levels. The ontology provides knowledge on the four areas of robot information, environment mapping, time information and workspace information. The environment mapping knowledge set includes functionality to describe areas using both geographical and landmark based positioning. Particularly useful in this ontology is the inclusion of uncertainty in robot and landmark positions, which assists greatly in modelling the dynamics of a SLAM problem. Although this ontology currently dominates the SLAM field (see figure ~\ref{fig::venn_diagram} - ontology for others) it does not possess the knowledge set required for multi-agent systems. The onto4MAT ontology however is the first attempt to design an ontology for multi-agent teaming systematically \cite{Onto4MAT}. This ontology enables an operator to provide an intent as tasks to a multi-agent system, and for the agents to then provide feedback to the operator.

\subsection{Sub-symbolic and Symbolic SLAM in Dynamic Environments}\label{sec::SLAM_Dynamic}
Spatial-temporal reasoning (as defined by the APA dictionary of Psychology) is \enquote{the ability to conceptualise the three dimensional relationships of objects in space and and to mentally manipulate them as a succession of transformations over a period of time.} This ability to conceptualise a 3D environment is vital for pose estimation (and as such environmental mapping). For SLAM methods that use image-to-image based tracking methods, some sort of spatial-temporal reasoning is required to maintain a map of the environment. Methods similar to that proposed in \cite{OrbSlam2} use ORBs as described above  to construct feature points to achieve this. This architecture sees the key frames tracked in a spanning tree, which allows for the direct mapping of map points on a co-visibility graph. However, this SLAM technique often fails to accurately map environments in dynamic scenarios that contain moving objects (e.g., people). Motion tracking is the ability of a system to be able to follow the motion of an object across multiple frames of a video. Deep Sort (in conjunction with YOLO)  is one of many algorithms currently available that provides this capability \cite{Wu2021SORTYMAA}. Motion tracking of dynamic objects in a scene has recently been combined with filtering techniques and Pose-Slam technology to alleviate the dynamic scenario problem alluded to above \cite{YOLO_Slam}, \cite{Yolo_Slam_2}, \cite{yolo_slam_3}, \cite{DS_SLAM}, \cite{Bescs2018DynaSLAMTM}, \cite{Ai2020DDLSLAMAR}, \cite{Fan2020SemanticSW}, \cite{Han2020DynamicSS}, \cite{Zhang2018SemanticSB}, \cite{Li2019StudyOS}. The common theme shared amongst all of these papers is that YOLO is used to detect objects and a low-level semantic understanding is used to determine whether they are dynamic. This is then used to determine whether to filter out parts of a frame, which is then fed into OrbSlam2 to allow for environmental mapping. The issue that plagues sub-symbolic SLAM algorithms in dynamic environments is less of a concern however, in architectures that utilise symbolic methods. Utilising landmark detection or similar with either pre-processing or an embedded module ensures that the system is much more resilient to dynamic feature interference. This is because the semantic label associated with each feature allows for the system to understand whether each feature is dynamic or static inherently.

\subsection{Simulated Environment} \label{sec::sim_env}
For use in this project, we assert that a simulated environment as a computer-based 3D platform capable of supporting a number of different sensor types launched on simulated robotic platforms to produce close-to-reality motion patterns. Many environments were considered for the deployment of the proposed architectures such as the DCIST env as shown in \cite{Kimera}, the Gazebo env \cite{Koenig2004DesignAU}, the Webots env \cite{Webots} and many others (see simulation on Venn diagram in figure ~\ref{fig::venn_diagram} for more). The EyeSim env \cite{braunl_2021} was selected for its hardware interoperability and support network through both documented resources and ongoing development support networks. This environment supports the deployment of many simulated robotic platforms in a joint space with visual, depth-based and IMU based sensor feeds. Agent communication is supported through radio and Wi-Fi based communications platforms, and the supported development languages are python and C / C++. The Unity-3D physics engine is the basis for the EyeSim environment, which incorporates error into the robotic movements. The intended hardware platforms (EyeBots) are described in \cite{braunl_2021} and the Unity-based system simulation environment aims to provide a realistic platform to match the actual EyeBot movements.

\subsection{Robotics} \label{sec::robotics}
The MAS SYMBO-SLAM architecture is to be deployed on both simulated and physical robotic platforms for experimental and testing purposes. The implementation on hardware will focus on unmanned ground vehicle (UGV) platforms capable of ground search tasks.  \cite{magic}, \cite{MAS_graph_slam} and \cite{semantic_map_matching_2022} utilise wheeled systems to achieve SLAM on a MAS. Other implementations include a system mounted to a car as seen in \cite{prob_semantic_SLAM} or \cite{Hierarchical_SLAM} and the use of aerial drones to conduct ground mapping observed in \cite{PGO1} or target localisation shown in \cite{localization_3D_moving_target}. Combinations and extensions of these implementation strategies, such as synchronous UAV and UGV deployment \cite{collab_semantic_understandng} are also possible (see figure ~\ref{fig::venn_diagram} - robotics for more implementation methods).

\section{Conclusion}\label{sec::conclusion}
This survey paper has extensively explored the integration of symbolic representation in Simultaneous Localization and Mapping (SLAM) within the context of multi-agent systems (MAS) and human-machine teaming. The systematic literature review, conducted following the PRISMA standard and software engineering guidelines, has identified a significant corpus of research that bridges the gap between symbolic reasoning, ontological design, and their application in SLAM tasks. The review highlights the burgeoning interest in deploying symbolic reasoning through ontological design to create more effective, context-aware maps in both centralized and collaborative agent systems. Advancements in artificial intelligence, particularly in edge computing and visual reasoning, have catalyzed the integration of symbolic and sub-symbolic reasoning, offering promising avenues for enhanced environmental mapping and situational awareness. This survey underscores the criticality of transparency in AI-enabled systems, especially in military applications, where the ethical use of such technology hinges on the operator’s ability to command and trust the system. The fusion of ontological knowledge with existing SLAM techniques presents a promising direction for future research, especially in creating dynamic, context-rich maps that enhance operator awareness in complex environments. The deployment of these advanced systems in simulated and real-world scenarios further demonstrates their potential applicability across various domains. This paper emphasizes the importance of continued exploration in this field, particularly in enhancing the trust architecture within human-machine teamed environments through improved system transparency. The potential of symbolic reasoning in SLAM, combined with the advancements in MAS and human-machine collaboration, paves the way for more robust, efficient, and transparent mapping systems, crucial for a wide range of applications, from military operations to autonomous navigation.

\bibliographystyle{IEEEtran}
\bibliography{references}

\end{document}